\documentclass[10pt,twocolumn,letterpaper]{article}

\usepackage[pagenumbers]{cvpr} 

\usepackage{graphicx}
\usepackage{amsmath}
\usepackage{amssymb}
\usepackage{booktabs}
\usepackage{makecell}
\usepackage{subcaption}
\usepackage{tikz}

\usepackage[pagebackref,breaklinks,colorlinks]{hyperref}

\usepackage[capitalize]{cleveref}
\crefname{section}{Sec.}{Secs.}
\Crefname{section}{Section}{Sections}
\Crefname{table}{Table}{Tables}
\crefname{table}{Tab.}{Tabs.}

\def\cvprPaperID{*****} 
\def\confName[pagenumbers]{CVPR}
\def\confYear{2022}

\begin{document}

\title{Sampling Strategies for Efficient Training of Deep Learning Object Detection Algorithms}

\author{Gefei Shen, Yung-Hong Sun, Yu Hen Hu, and Hongrui Jiang\\
University of Wisconsin - Madison\\
Department of Electrical and Computer Engineering \\
Madison, WI 53706\\
{\tt\small gshen22@wisc.edu, ysun376@wisc.edu, yhhu@wisc.edu, hongruijiang@wisc.edu }
}
\maketitle

\begin{abstract}
Two sampling strategies are investigated to enhance efficiency in training a deep learning object detection model. These sampling strategies are employed under the assumption of Lipschitz continuity of deep learning models. The first strategy is uniform sampling which  seeks to obtain samples evenly yet randomly through the state space of the object dynamics. The second strategy of frame difference sampling is developed to explore the temporal redundancy among successive frames in a video. Experiment result indicates that these proposed sampling strategies provide a dataset that yields good training performance while requiring relatively few manually labelled samples.
\end{abstract}

\section{Introduction}

Sun et al.'s \cite{sun2025easyvis2realtimemultiview} laparoscopy project demonstrates an informative way of data collection and labeling. The project begins with a small set of initial training data, and it iteratively improves the model through a semi-automated data labeling pipeline. Initially a small batch of training data is used to train the model. Then the model is used to make predictions while a human annotator observes its real time prediction result. The annotator then records the frame where the model prediction is off, and manually correct these predictions. Corrected samples are then used to improve the model. This iterative process aims to find the cases where the model's performance is sub-standard and strengthen the model's performance on these cases.

The selective sampling approach taken by Sun et al.\cite{sun2025easyvis2realtimemultiview} can be characterized as a "human based" active learning. The selection of additional training data is based on human annotator's observation and decision. However, to scale this process to enhance efficiency, an automated active learning strategy is required. 

In this work, we builds on this manual active learning pipeline to explore two specific sampling strategies: uniform sampling and frame difference sampling. Specifically, we employ a pre-trained YOLOv8 model and customize it to detect and estimate the poses of laparoscopic surgical tools. 

\section{Background}

\subsection{Sampling and Active Learning}
Machine learning (ML) models need to be trained with data samples. The quality of trained ML models often is tied to the quantity (number of training samples) and quality (accuracy of labels) of the training data. Insufficient amount of training samples may yield inferior ML models that do not generalize well. On the other hand, acquiring samples and labeling samples both may consume significant resources. Therefore, it is desirable to judiciously acquire sample and label samples to achieve desired performance of trained ML models while minimizing the sampling and labeling cost. 

Experiment design \cite{kirk2009experimental} may be applied initially to setup the sampling strategy. However, it is often unclear how many training samples are required to achieve desired performance by testing a trained ML model against a preset {\it validation} dataset. If the sampling (and labeling) cost is of no concern, one would use as many training samples as practically possible. However, if the sampling and/or labeling cost is a constraint, an active learning \cite{Settles_2009} approach may be desired. With active learning, an ML model is trained incrementally. At each iteration, the active learning algorithm will select a subset of unused samples and query their labels from an {\it orcle} which may be a human labeler. The selection criteria may depend on how well the current ML model is developed. 

In this work, we will investigate sampling techniques for a laparoscopy project\cite{sun2025easyvis2realtimemultiview}. In this project, a ML model YoloV8-detect \cite{yolov8_ultralytics} will be trained with images extracted from videos acquired in the laboratory and manually labeled. The labeling cost in terms of the numbers of images that need to be manually labeled is to be minimized.  

\subsection{Lipschitz Continuity of ML Models}

Many ML models may be approximated by a higher-dimension {\it Lipschitz} function $f: \mathbb{R}^2 \rightarrow \mathbb{R}^n$ \cite{virmaux2018lipschitz}. Denote $x$ and $y$ to be the data and label respectively. The Lipschitz condition stipulates that if for some distance metric $\delta$,

\begin{equation}
    \delta(x_1, x_2) < \sigma
\end{equation}

The corresponding ML model $f(\cdot)$ will satisfy
\begin{equation}
    \delta (f(x_1), f(x_2)) < \epsilon(\sigma)
\end{equation}
where $f(x_1)$, $f(x_2)$ are the corresponding labels of $x_1$ and $x_2$, and $\epsilon$ is a function.

In machine learning, imposing Lipschitz continuity to the loss function during training ensures the trained machine learning model will not be overly sensitive to small perturbations in the input, leading to more robust, stable, and predictable behavior. leveraging Lipschitz continuity, active learning algorithms can select more informative data points for labeling, leading to more accurate and reliable models. 

\subsection{Sampling Strategies}

\subsubsection{State Space Sampling}
In this work, each sample is a frame of video (an image). If we rasterize the pixel values into a vector, it would have high dimension. Direct sampling in such a high-dimensional feature space would be very inefficient. However, by performing experiments using the prototype system, we can produce training video under different scenarios. To be more specific, the goal of object detection in this application is to estimate the 3D positions and orientations of surgical tools. These are the {\it states} of objects that need to be estimated. Hence, it is better to sample in the contextual {\it state space} rather than the context-agnostic pixel value space. Therefore, in this work, we perform experiments that covers all the operating scenarios with the prototype laparoscopic surgical system. These include different positions, orientations of the surgical tools in the training box. The objective is to generate training videos that span the state space in which the object's state is to be estimated. 

\subsubsection{Uniform Sampling}

Uniform sampling attempt to randomly sample data from the feature distribution according to a uniform distribution. Uniform sampling promises that samples may cover the entire feature space. In this work, we apply uniform sampling in the state space to boot-strap the active learning. A ML model trained with these uniformly sampled training data is likely to provide a crude sketch of the final model over the contextual relevant sampling space.

The granularity of this initial boot-strapping uniform sampling is determined by the amount of samples that will be labeled (sampling cost) for this purpose. It also depends on the Lipschitz continuity property of the loss function for training the ML model.

\subsubsection{Frame Difference Sampling}

The other utilization of the Lipschitz continuity property is in the temporal domain. It is observed that high correlation exists between successive video frames (temporal Lipschitz continuity), we examine magnitude of frame difference to identify most informative video frames for the purpose of sampling. 

Specifically, we convert the video frames into grayscale and compute the frame by frame difference.

\begin{equation}
	D_t = \sum_{u,v}\bigl|I_t(u,v) - I_{t-1}(u,v)\bigr|
\end{equation}

Where $I_t(u,v)$ is the intensity of pixel $(u,v)$ in frame $t$. Then we rank all frames by $D_t$ and select the top $P$ of frames for manual labeling. This approach ensures that our training set is consist of with the most dynamic and high-variance samples.

\section{Methods}

In this work, we hypothesize that:

\begin{enumerate}
    \item Uniform Sampling: Using the original video sequence, we select $P$ frames evenly from the video sequence. Spacing samples evenly in time should ensure broad coverage of all video segments. We expect that even without targeting specific complex samples, the model would steadily improve as it sees representative samples from across the entire video sequence.
    \item Frame Difference sampling: We hypothesize that frames with large pixel changes represents more complex samples like rapid motion, change in light conditions that the model is likely to find it challenging.
\end{enumerate}

Both hypothesized methods should achieve better performance in the Sun et al.'s\cite{sun2025easyvis2realtimemultiview} applications.

\subsection{Experiment Setup}
In this work, our plan is to train a deep neural network object detection algorithm YoLO v8 to learn new objects. The model used is pre-tained YoLo v8-n models release by Ultralytics trained on COCO datasets \cite{yolov8_ultralytics}. Our goal is to fine-tune this pre-trained YoLO model using videos captured using the on-board micro-camera array of a prototype laparoscopic surgical visualization system EasyViz. Some captured video frames of the object (a surgical grasper) are shown in Fig. \ref{grasper picture}. The outcome of object detection is the coordinates of a bounding box that tightly encloses the object. Hence, this is a regression problem. 

\subsubsection{Dataset\label{sec:dataset}}

The video sequence used to finetune the model is a 140 frame video sequence captured at 30hz of a grasper manipulating a bean. We also produced a dataset from the same setup that includes mostly challenging corner cases to evaluate the finetuned model's performance.

For this evaluation dataset. First, we sample the video frames with variable intervals. This is because the object motion is much slower than the video frame rate of 30 Hz. Hence, the content of successive video frames are highly correlated and redundant. M0 is used to predict the object bounding box coordinates for each sub-sampled video frame. The prediction results, denoted by R0 will be examined by a human operator in a visualized form \ref{visualization picture}. And if the predicted bounding box position is visibly incorrect, the correct coordinates will be manually entered. Otherwise, the predicted coordinates will be considered as the label of that video frame. This step yields 797 labeled images of the objects.

\begin{figure}
	\centering
	\includegraphics[width=1\linewidth]{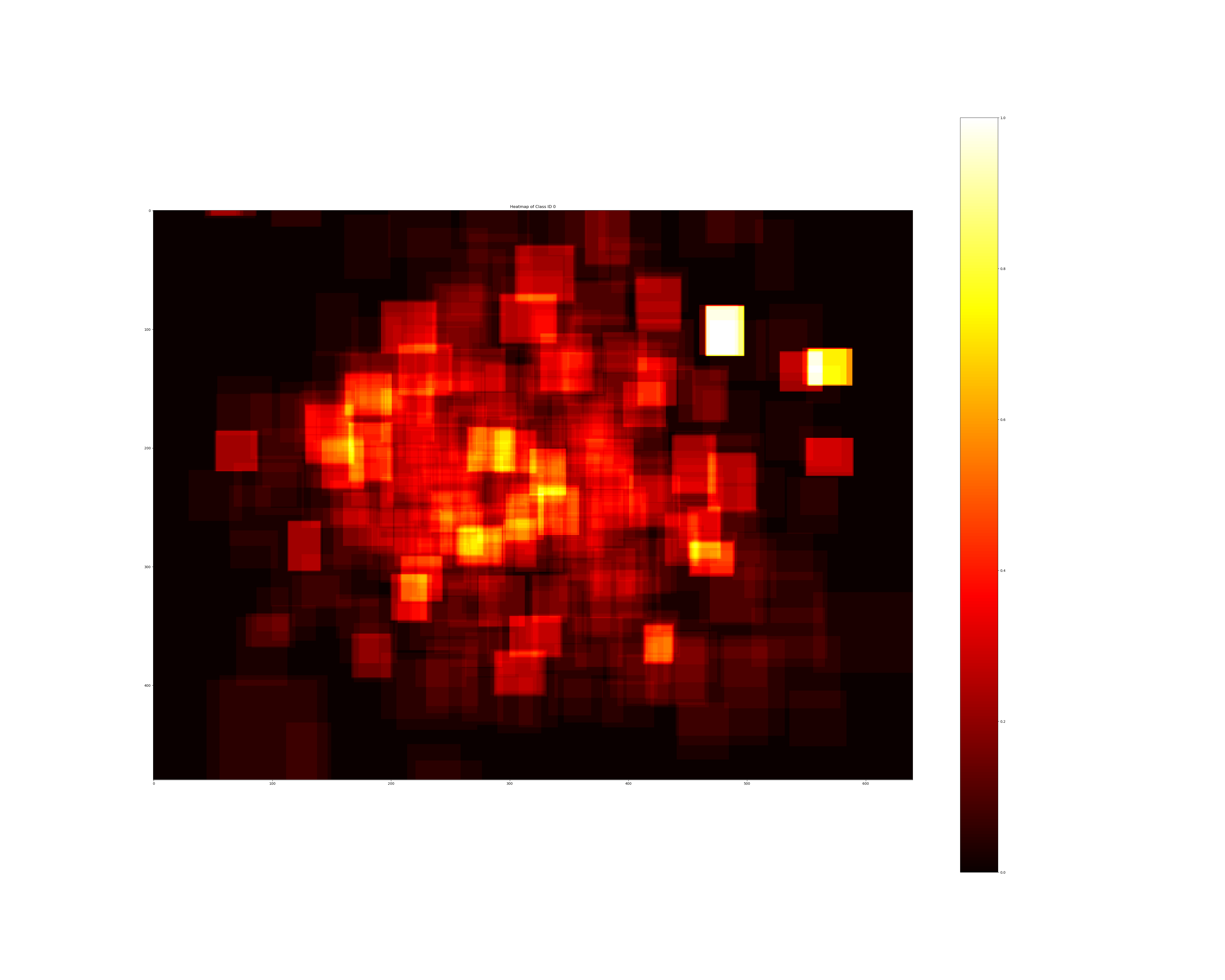}
	\caption{Bean Heatmap}
	\label{fig:bean heatmap}
\end{figure}

\begin{figure}
	\centering
	\includegraphics[width=1\linewidth]{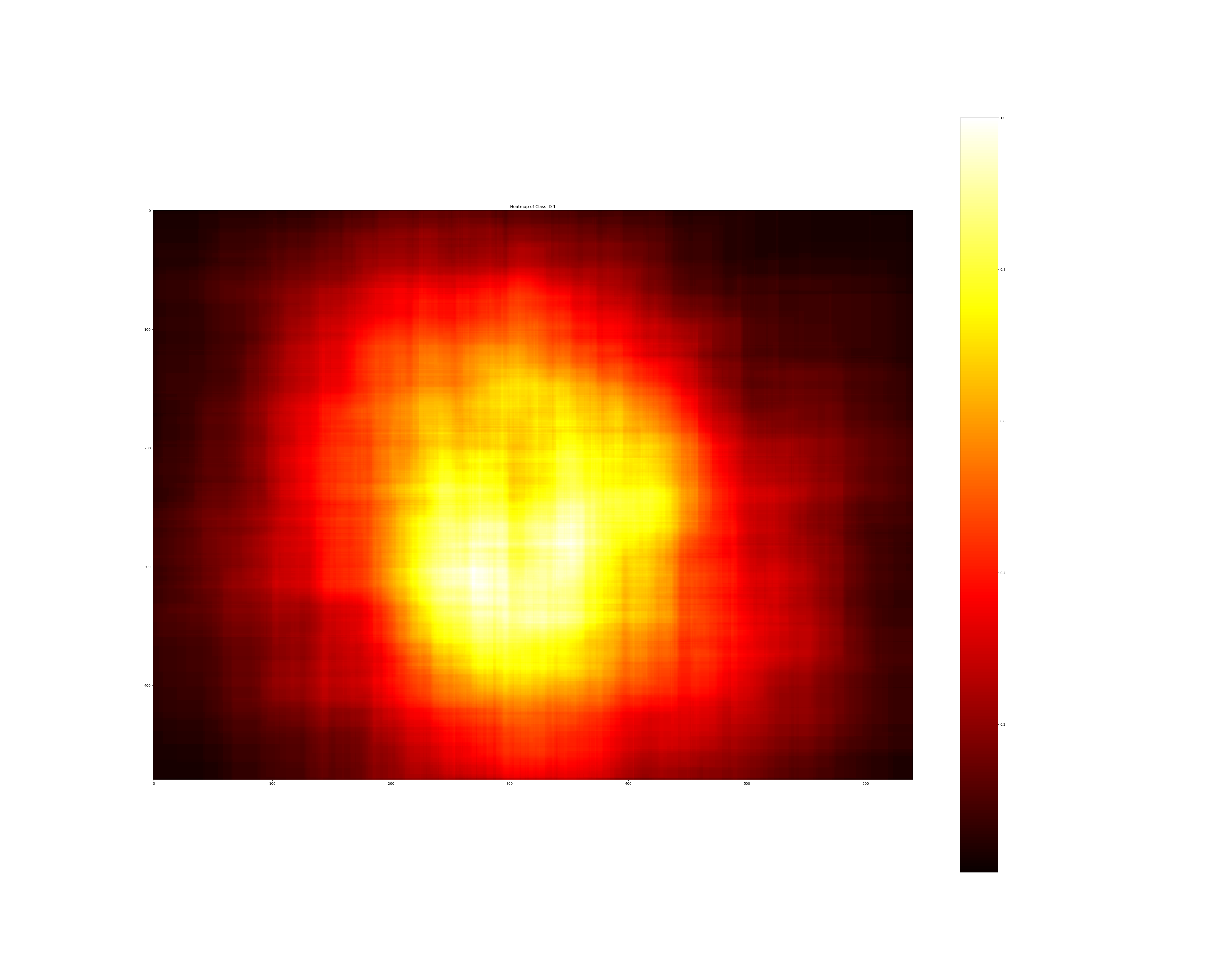}
	\caption{Grasper Heatmap}
	\label{fig:grasper heatmap}
\end{figure}

The dataset includes two types of objects, grapser and bean. For our experiment, we focus mainly on the performance of the grasper. Fig\ref{fig:bean heatmap} and \ref{fig:grasper heatmap} show the heat maps of both classes and the occurrences of both classes are listed below:

\begin{itemize}
	\item Image with bean(s): 346
	\item Image with grasper(s): 762
	\item Image with bean(s) only: 34
	\item Image with grasper(s) only: 450
	\item Image with both bean(s) and grasper(s): 312
\end{itemize}

\begin{figure}
	\centering
	\includegraphics[width=1\linewidth]{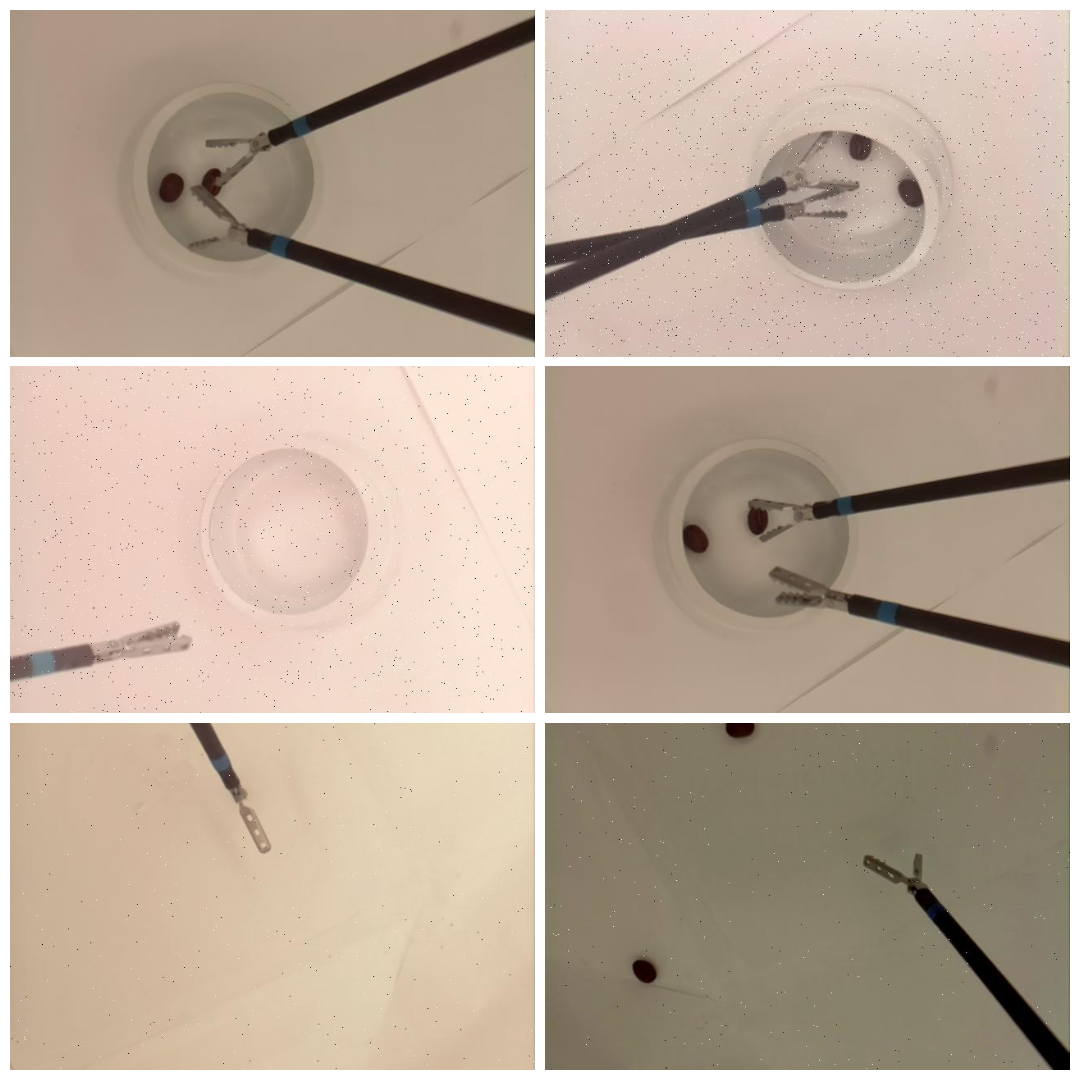}
	\caption{Testing Dataset Sample}
	\label{fig:grasper picture}
\end{figure}

\subsubsection{Experiment Steps}

We evaluate two sampling strategies (uniform and frame difference) across five training-set sizes (50\%, 33.3\%, 20\%, 10\%, 5\%).  For each combination, we fine-tune the same pretrained YOLOv8n model and measure performance on both the held-out frames from the original video and on an entirely separate, non-consecutive video dataset.

\begin{enumerate}
	\item \textit{Define training percentages.}  
	Let $P \in \{50\%, 33.3\%,\,20\%,\,10\%,\,5\%\}$ be the fraction of frames used for fine-tuning.
	
	\item \textit{Uniform sampling.}  
	For each $P$:  
	\begin{itemize}
		\item Sort frames in temporal order and select every $\bigl\lfloor 1/P \bigr\rfloor$-th frame to achieve approximately $P$ coverage.  
		\item Fine-tune the pretrained YOLOv8n for 20 epochs on this subset, yielding model $M_{\mathrm{uni},P}$.
	\end{itemize}
	
	\item \textit{Frame difference sampling.}  
	For each $P$:  
	\begin{itemize}
		\item Compute grayscale difference $D_t = \sum_{u,v}\bigl|I_t(u,v) - I_{t-1}(u,v)\bigr|$ for each frame $t$.  
		\item Select the top $P$ fraction of frames with the largest $D_t$ values.  
		\item Fine-tune YOLOv8n for 20 epochs on this subset, yielding model $M_{\mathrm{diff},P}$.
	\end{itemize}
	
\end{enumerate}

\subsection{Evaluation Metrics}
For every trained model $M_{s,P}$, we run inference on left out video frames in the training video sequence and record per-class intersection over union (IoU) to evaluate the model's prediction stability. A empirical analysis of Lipschitz constant is also calculated using:
\begin{equation}
	{K}_{percent}^{s,P}
	\;=\;
	\operatorname{Quantile}_{percent}
	\Bigl\{
	\frac{\lvert f(x_i)-f(x_j)\rvert}{\lVert x_i-x_j\rVert}
	\Bigr\}_{1\le i<j\le N}
\end{equation}

The model's performance is also evaluated on the testing dataset, we report precision, recall and mAP@0.5 to evaluate model's performance on real world data.

\section{Result}

\subsection{Stability of Model Prediction}

\begin{figure*}[t]
	\centering
	\renewcommand{\thesubfigure}{\arabic{subfigure}}
	\setcounter{subfigure}{0}
	
	\begin{subfigure}[b]{0.48\textwidth}
		\centering
		\includegraphics[width=0.8\linewidth]{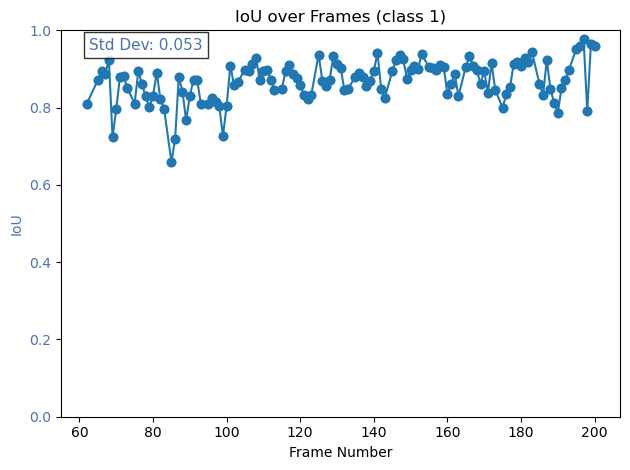}
		\caption{Uniform\,10\%}
	\end{subfigure}\hfill
	\begin{subfigure}[b]{0.48\textwidth}
		\centering
		\includegraphics[width=0.8\linewidth]{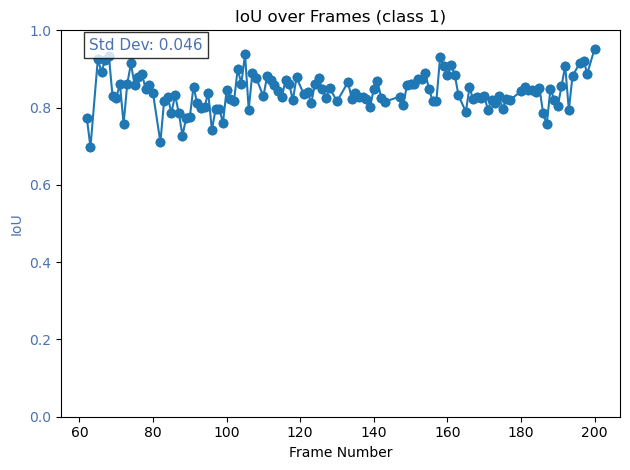}
		\caption{FrameDiff\,10\%}
	\end{subfigure}
	
	\vspace{0.1em}
	
	\begin{subfigure}[b]{0.48\textwidth}
		\centering
		\includegraphics[width=0.8\linewidth]{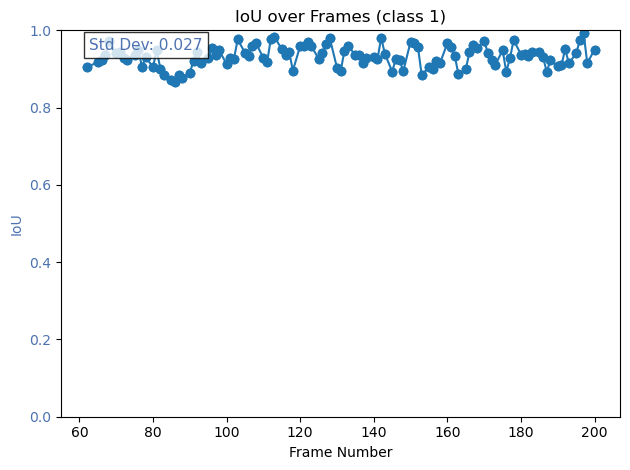}
		\caption{Uniform\,20\%}
	\end{subfigure}\hfill
	\begin{subfigure}[b]{0.48\textwidth}
		\centering
		\includegraphics[width=0.8\linewidth]{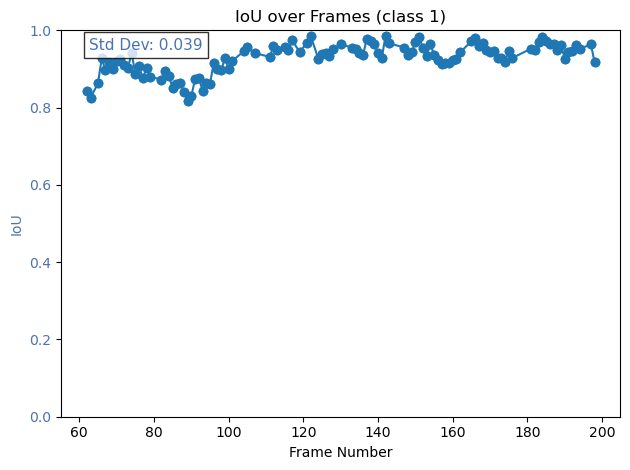}
		\caption{FrameDiff\,20\%}
	\end{subfigure}
	
	\vspace{0.1em}
	
	\begin{subfigure}[b]{0.48\textwidth}
		\centering
		\includegraphics[width=0.8\linewidth]{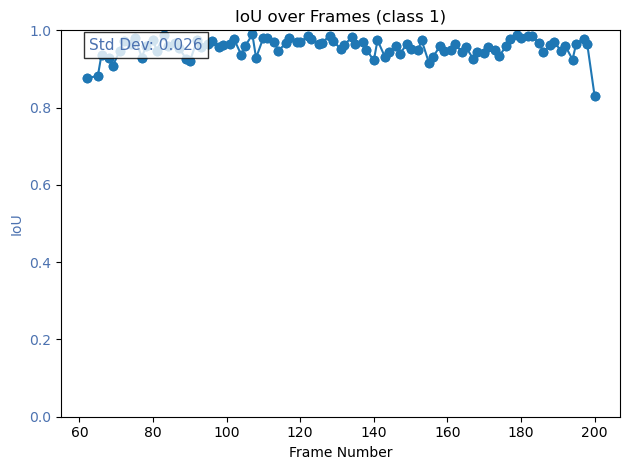}
		\caption{Uniform\,33\%}
	\end{subfigure}\hfill
	\begin{subfigure}[b]{0.48\textwidth}
		\centering
		\includegraphics[width=0.8\linewidth]{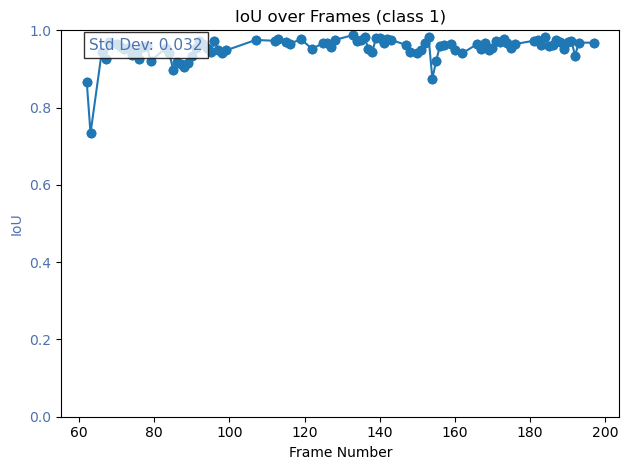}
		\caption{FrameDiff\,33\%}
	\end{subfigure}
	
	\vspace{0.1em}
	
	\begin{subfigure}[b]{0.48\textwidth}
		\centering
		\includegraphics[width=0.8\linewidth]{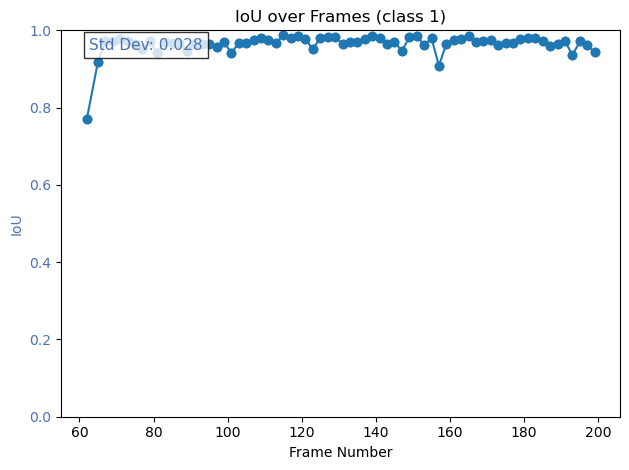}
		\caption{Uniform\,50\%}
	\end{subfigure}\hfill
	\begin{subfigure}[b]{0.48\textwidth}
		\centering
		\includegraphics[width=0.8\linewidth]{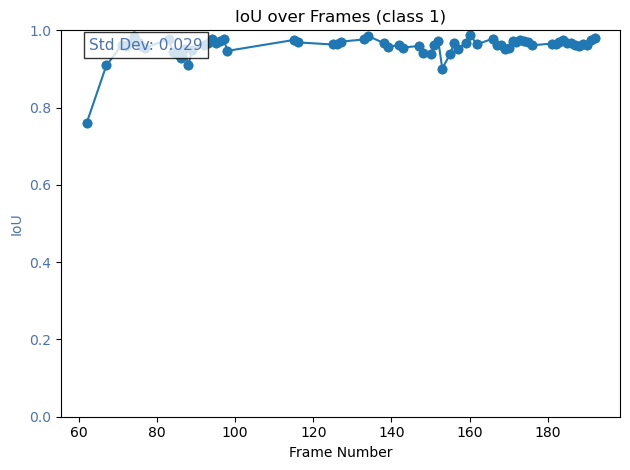}
		\caption{FrameDiff\,50\%}
	\end{subfigure}
    
	\caption{IoU stability over consecutive frames for each sampling strategy and budget.  
		Each row shows the same label budget ($P$). Left row: Uniform Sampling; Right row: Frame Difference Sampling}
	\label{fig:iou_stability}
\end{figure*}

Figure \ref{fig:iou_stability} visualizes the per frame IoU for both sampling strategies at four budget levels. As the budget increases, the model produces smoother and more consistent IoU curves. By contrast, a lower budget cause the model to produce a less consistent IoU curve. 

We quantify these observations using the table\ref{tab:lipschitz_grasper}, which reports empirical Lipschitz constant percentiles ($K_{50}$, $K_{90}$, $K_{95}$, $K_{99}$) for the grasper class. 

\begin{table}[ht]
	\centering
	\begin{tabular}{|c|c|c|c|c|c|}
		\hline
		\textbf{Strategy} & \textbf{$P$ (\%)} & $\boldsymbol{K_{50}}$ & $\boldsymbol{K_{90}}$ & $\boldsymbol{K_{95}}$ & $\boldsymbol{K_{99}}$ \\
		\hline
		Uniform & 10 & 0.499 & 1.769 & 2.367 & 7.280 \\
		Uniform & 20 & 0.279 & 1.340 & 1.926 & 4.825 \\
		Uniform & 33 & 0.136 & 0.428 & 0.600 & 1.506 \\
		Uniform & 50 & 0.101 & 0.359 & 0.531 & 1.284 \\
		\hline
		Frame Diff & 10 & 0.658 & 1.957 & 3.096 & 8.586 \\
		Frame Diff & 20 & 0.202 & 0.579 & 0.880 & 2.857 \\
		Frame Diff & 33 & 0.139 & 0.407 & 0.671 & 2.632 \\
		Frame Diff & 50 & 0.094 & 0.417 & 0.730 & 2.554 \\
		\hline
	\end{tabular}
	\caption{Empirical Lipschitz constant percentiles ($K_{50}$, $K_{90}$, $K_{95}$, $K_{99}$) for the \textit{grasper} class under each sampling strategy and label budget.  Lower values indicate more stable predictions across consecutive frames.}
	\label{tab:lipschitz_grasper}
\end{table}

The Lipschitz constant back up the visual observations in Figure \ref{fig:iou_stability}. As the labeling budget $P$ increases, for both strategy, their $K_{50}$, $K_{90}$ and $K_{90}$ drops steadily, meaning the model's frame to frame prediction is getting more consistent. However, the $K_{99}$ reading exhibits a different behavior. As the labeling budget increases, uniform sampling strategy keeps reducing the rare jumps in IoU. On the other hand, frame difference strategy did not successfully improve the $K_{99}$ reading.

This suggests that even when we have a large labeling budget (for example, 50\%), some abrupt change in scene can still trigger large frame to frame IoU fluctuations under the frame difference sampling policy. Whereas the uniform sampling is likely to smooth out those fluctuations. This characteristic makes the frame difference sampling strategy not ideal for the laparoscopy project, where the project requires temporal continuity.

\subsection{Detection Accuracy}

Table~\ref{tab:sampling_test_results} below reports precision, recall, and $\mathrm{mAP}@0.5$ for the two sampling strategies at four training budgets ($P=10,20,33,50\,\%$).  
The numbers refer to performance on the independent corner case testing dataset described in Section~\ref{sec:dataset}, results for a naive random sampler (average of file runs) are included for reference.

\begin{table}[ht]
	\centering
	\begin{tabular}{|c|c|c|c|c|}
		\hline
		\textbf{Strategy} & \textbf{P (\%)} & \textbf{Precision} & \textbf{Recall} & \textbf{mAP@0.5} \\
		\hline
		Uniform & 10 & 0.00729 & 0.85224 & 0.29126 \\
		Uniform & 20 & 0.00795 & 0.88648 & 0.33151 \\
		Uniform & 33 & 0.97957 & 0.10000 & 0.51235 \\
		Uniform & 50 & 0.84309 & 0.44429 & 0.67144 \\
		\hline
		Frame Diff & 10 & 0.00735 & 0.84698 & 0.25539 \\
		Frame Diff & 20 & 0.00747 & 0.86173 & 0.25396 \\
		Frame Diff & 33 & 0.98921 & 0.05047 & 0.32841 \\
		Frame Diff & 50 & 0.82772 & 0.62004 & 0.72505 \\
		\hline
        Random & 10 & 0.007002 & 0.828436 & 0.224504 \\
        Random & 20 & 0.00706 & 0.82662 & 0.26307 \\
        Random & 33 & 0.00783 & 0.87945 & 0.29975 \\
        Random & 50 & 0.95504 & 0.10416 & 0.40243 \\
        \hline
	\end{tabular}
	\caption{Comparison of Sampling Strategies at Varying Data Percentages (Testing Dataset)}
	\label{tab:sampling_test_results}
\end{table}

\subsubsection{Low Budget}
With at most 28 labeled frames, all samplers achieve high recall ($\approx0.83-0.89$) but almost zero precision ($\approx0.007$), yielding unoptimal $\mathrm{mAP}$ values (0.22-0.33). Suggesting that COCO priors dominate when the finetuning set is extremely sparse. Among the three choices, uniform sampling still edges out the others, presumably because it guarantees at least one example from every temporal segment, while the difference heuristic concentrates on high motion snippets and neglects static backgrounds.

\subsubsection{Medium Budget}
At roughly 33\% of the video sequence, uniform sampling now delivers $0.98$ precision and doubles recall relative to frame difference (0.10 vs.\ 0.05), producing a 56\% higher $\mathrm{mAP}$ (0.512 vs.\ 0.328). Seeing a balanced cross section of the video helps the model suppress false positives without sacrificing all true detections, whereas the motion biased subset still lacks many typical poses. Random sampling is the weakest, indicating that naive sampling cannot utilize the larger budget.

\subsubsection{High Budget}
Once half of the frames are annotated (70 labels), frame difference overtakes uniform and attains the best overall detector: $\mathrm{mAP}=0.725$, precision $0.828$, recall $0.620$. This is due to the additional high motion samples supply rare viewpoints and motion blur patterns that uniform undersamples. Random sampling's precision jumps to 0.96 but recall collapses to 0.1, resulting in a low $\mathrm{mAP}$ of 0.4, which is far below the other two.

\subsubsection{Summary}
Uniform sampling improves model performance proportionally as $P$ grows, which can be a safe choice when the budget is low. It provides a broader coverage to the temporal video sequence. Once there is baseline coverage, the model would benefit more with corner case samples from frame difference sampling. By contrast, random sampling offers no systematic advantage at any budget.

\section{Conclusion}

This work presents a end to end active learning pipeline for low cost adaptation of a pretrained Yolov8 model to a laparoscopic training scenario. Two sampling heuristics were evaluated. When the sampling budget is small, a uniform sampling provides a bottom line safety guarantee for generalization. After the model generalizes well, a frame difference sampling strategy can sample corner cases and further improve model performance.

\subsection{Limitation and Future Work}
\begin{itemize}
	\item \textbf{Single Video Training}
		Experiment is conducted on a short single video recording. The model's generalization to other applications is untested.
	\item \textbf{Detection Task Only}
		The pipeline only works for 2D detection tasks, it's application to more complex tasks like pose estimation is unkown.
\end{itemize}

In future work we plan to repeat the study with multiple video recordings and build a hybrid active learning strategy based on the current two heuristics.

\section{Acknowledgement}
This work was supported by the National Institute of Biomedical Imaging and Bioengineering (NIBIB) of the U.S. National Institutes of Health (NIH) under award number R01EB019460.


{\small
\bibliographystyle{ieee_fullname}
\bibliography{egbib}
}

\end{document}